\documentclass[10pt, a4paper]{article}
\usepackage{lrec2006}
\usepackage[T1]{fontenc}
\usepackage[utf8]{inputenc}
\usepackage{times}
\usepackage{latexsym}
\usepackage{url}
\usepackage{rotate}
\usepackage{fancyvrb}
\usepackage{xcolor}
\usepackage{multirow}
\usepackage{graphicx}

\newcommand{\levA}[1]{{\bf #1}}
\newcommand{\levB}[1]{ ~~{#1}}
\newcommand{\levC}[1]{ ~~~~~~{#1}}
\newcommand{\ex}[1]{\emph{#1}}
\newcommand{\m}[1]{[\textbf{#1}]}
\newcommand{\ra}{\ensuremath{\rightarrow}}

\title{Mining Naturally-occurring Corrections and Paraphrases \\ from Wikipedia's Revision History}

\name{Aurélien Max\hspace{0.5cm} Guillaume Wisniewski}

\address{LIMSI-CNRS and Universit\'e Paris-Sud 11 \\
  Orsay, France\\
  {\tt \{aurelien.max,guillaume.wisniewski\}@limsi.fr}
}

\abstract{Naturally-occurring instances of linguistic phenomena are
  important both for training and for evaluating automatic processes
  on text. When available in large quantities, they also prove
  interesting material for linguistic studies.  In this article, we
  present a new resource built from Wikipedia's revision history,
  called WiCoPaCo (Wikipedia Correction and Paraphrase Corpus), which
  contains numerous editings by human contributors, including various
  corrections and rewritings. We discuss the main motivations for
  building such a resource, describe how it was built and present
  initial applications on French.}

\begin{document}

\maketitleabstract

\section{Introduction}
\label{section:introduction}

This paper describes the construction of a corpus of rewritings
extracted from the revision history of Wikipedia, which includes
spelling corrections, reformulations, and other local text
transformations. Such rewritings are of interest for many NLP
applications, including text correction and normalization,
paraphrasing, summarization, etc. For many of these applications, only
a few hand-crafted or artificial corpora of small size are available,
which prevents researchers from using machine learning techniques
requiring important amounts of training examples and questions the
validity of evaluations that use them. For instance, the study
reported in~\cite{Schroeder_09_EACL} shows the negative impact of
using artificially produced sentential paraphrases in a multi-source
Machine Translation experiment.

While the cost of the annotation effort has always been a burden to
the creation of huge corpora of naturally-occurring rewritings, we
believe that the growth of publicly editable wikis with high
contribution rates allows us to easily collect large amounts of useful
rewriting examples. Indeed, an important characteristics of Wikipedia
(and other wikis) is the fact that users not only contribute new
content but also improve the overall quality of the text collection
(an encyclopedia in the case of Wikipedia), counteracting spam and
making various types of corrections and improvements to the created
texts.

The huge amounts of quality data in Wikipedia have triggered many
works on automatic resource acquisition (e.g. acquisition of
lexical-semantic knowledge \cite{Zesch_08_LREC}). Closer to our work,
\cite{Nelken_08_AAAI} exploit Wikipedia's revision history for
acquiring instances of \emph{eggcorns} (semantically plausible
homophonic confusions) and their correction, as well as text spans and
their compressed rewritings.  These correspondences are found by
applying a search for longest common subsequences (using the same
algorithm as the \texttt{diff} command) between any two consecutive
versions of articles and identifying substitutions in the results. In
their work, a very simplifying assumption is made that such pairs
correspond to instances of text compressions whenever the rewritten
text is shorter than the original text.

There is, however, a much greater variety of rewritings
that are also of great interest for several NLP applications. In this
work, we describe the construction of a new resource from Wikipedia's
revision history. The raw resource contains all types of local
rewritings found in the encyclopedia, with their context and various
meta-data related to them. Independent annotation efforts can then
assign labels to the data depending on a targeted application,
yielding data suitable for supervised machine learning and for
evaluation of NLP tools.  Furthermore, the collected data constitutes
a huge collection of naturally-occurring corrections and rephrasings
of particular interest for writing studies.

\section{Building the resource}

In this section, we describe the main details behind the construction
of the WiCoPaCo corpus\footnote{WiCoPaCo can be freely downloaded from
  the website \url{http://wicopaco.limsi.fr}} (Wikipedia Correction
and Paraphrase Corpus). Our objective is to build a resource suitable
for many types of studies on naturally-occurring local
rewritings. However, we must set some practical bounds as not all
rewritings are of equal interest.

The construction of the corpus is in done in two steps. In a first step, a set
of local modifications is extracted by computing, from a Wikipedia dump stored
in a local database, the differences among any two versions of all articles
using the efficient longest common subsequence algorithm. All paragraphs
containing at least one substitution are extracted, and their text is
normalized (de-wikification, tokenization, etc.) As the aim is to extract
local modifications, only rewriting implying at most 7 words are taken into
account.

This first step allows us to extract a very large number of local
modifications. Note that we do not consider modifications that involve
only additions or deletions of tokens, as this corpus is designed to
support the study of text modifications where two text spans in
context can be paired.

In a second step, we apply a set of hand-crafted filtering rules. In
particular, we filter out modifications in which the ratio of common
words in the original and the modified sentence (defined in a greedy
sense) is under a given threshold and changes that concern only
punctuation or case modifications. The first rule filters out
modifications that may significantly change the meaning of the
enclosing text unit. The second filter limits the size of the corpus,
although these occurrences can be kept for studies requiring them, as for
example studies on text ponctuation.

We record the full paragraph in which a local rewriting is found in
order to allow application to exploit a larger context than that of
the enclosing sentence.\footnote{In addition to the shorter context,
  sentences are not good contextual units for this work as sentence
  segmentation is difficult to define, even on encyclopedic
  text. Taking paragraphs as units allows making use of unambiguous
  elements, such as blank lines or structural boundaries. In any case,
  all the necessary metadata that link to the original Wikipedia
  article and revision are kept.} Because contributors can make
several edits when submitting a new revision, we record both the
context of the original phrase and that of the modified phrase. We
also record, as meta-data on the revision, all necessary identifiers
from the Wikipedia database, including the identifier of the user who
submitted the revision in which the text substitution was found and
the number of revisions made by this user. We purposefully do not
include revisions that were submitted by automatic bots, as we want to
restrict the data to modifications that could be made by human
contributors.\footnote{This, of course, does not mean that
  modifications programmed in bots are of no interest. Other reasons
  for not keeping them include the large quantities of modifications
  and their impact on modification frequency, the fact that bots can
  hardly take linguistic context into account and can introduce errors
  when their programmers did not anticipate cases where the bot's
  modifications should not apply.} Anonymous and registered human
users are distinguished, permitting further data mining to assess the
reliability of a given local substitution based on its author's
reputation \cite{Adler_08_WikiSym}.

An output XML file is finally produced with unique modification
identifiers, which can be used in subsequent work to associate
annotations to every instance.\footnote{At the time of writing, the
  resource contains 408,816 modifications in context.}
Figure~\ref{fig:ex_WiCoPaCo} shows an example of an entry of the
WiCoPaCo corpus. Our initial work was carried out on the French
version of the Wikipedia database.
Figure \ref{figure:typology} reports the main types of text
substitutions that are found by manual inspection of the corpus,
including many types which were not considered in
\cite{Nelken_08_AAAI}. We distinguished two main classes, namely that
of modifications where the original and the modified text convey
essentially the same meaning in context (according to human judgment),
and that of modifications where meaning has been modified for various
possible reasons.\footnote{Note that spam applied to our defined local
  modifications appears in this category.} Automatic classification of
modifications, using the subclasses of Figure~\ref{figure:typology} or
any application-oriented classes is part of our future work. For
instance, the automatic detection of what is referred to as ``subtle
grammatical spamming'' may be of great use for Wikipedia
administrators to locate incorrect changes that pollute the textual
database and might remain in the encyclopedia for long times before
they are spotted and corrected.

\begin{figure*}[htbp]
  \centering
\begin{Verbatim}[commandchars=@\[\],frame=single,fontsize=\small]
@PYaZ[<modif] @PYaQ[id=]@PYaB["23"] @PYaQ[wp_page_id=]@PYaB["7"] @PYaQ[wp_before_rev_id=]@PYaB["4649540"]
       @PYaQ[wp_after_rev_id=]@PYaB["4671967"] @PYaQ[wp_user_id=]@PYaB["0"]
       @PYaQ[wp_user_num_modif=]@PYaB["1096911"] @PYaQ[wp_comment=]@PYaB["Définition"]@PYaZ[>]
  @PYaZ[<before]@PYaZ[>]On nomme @PYaZ[<m] @PYaQ[num_words=]@PYaB["1"]@PYaZ[>]Algebre@PYaZ[</m>] linéaire la branche
  des mathématiques qui se penche...@PYaZ[</before>]
  @PYaZ[<after]@PYaZ[>]On nomme @PYaZ[<m] @PYaQ[num_words=]@PYaB["1"]@PYaZ[>]Algèbre@PYaZ[</m>] linéaire la branche
  des mathématiques qui se penche...@PYaZ[</after>]
@PYaZ[</modif>]
\end{Verbatim}
\caption{Sample XML entry of WiCoPaCo}
  \label{fig:ex_WiCoPaCo}
\end{figure*}

\begin{figure*}[!htpb]
\begin{center}
\begin{small}
\begin{tabular}{|p{5.5cm}|p{11cm}|}
\hline
\levA{Same meaning} & \\ \hline

\levB{Different spelling} & \\
\levC{Encyclopedic normalizations} & \ex{\m{Son 2ème disque --> Son deuxième disque}} \\
\levC{Unknown words due to spelling} & \ex{c' est-à- dire la \m{dernrière --> dernière} année avant l' ère chrétienne} \\
\levC{Missing diacritics} &  \ex{la jeune Natascha Kampusch ,\m{agée --> âgée} de 18 ans} \\
\levC{Homophonic confusions} & \ex{L' immense majorité de \m{ses --> ces} nobles vit dans des conditions} \\
\levC{Grammatical errors} & \ex{dans le but de \m{sensibilisé --> sensibiliser} sur les changements} \\ \hline

\levB{Different wording} & \\
\levC{Syntactic rewriting} & \ex{Le tritium \m{existe dans la nature . Il est produit --> se forme naturellement} dans l' atmosphère} \\
\levC{Paraphrases} & \ex{'Gimme Gimme Gimme' et 'I Have A Dream' \m{contribueront au gigantesque succès de --> viendront alimenter la gloire que connait} Abba} \\
\levC{Translation} & \ex{Bertrand Russell , dans \m{History of the Western Philosophy --> Histoire de la philosophie occidentale}} \\ \hline \hline

\levA{Different meaning} & \\ \hline

\levB{Acceptable meaning changes} & \\
\levC{Precision of meaning} & \ex{alors \m{que l' ordinateur -> qu'un processeur de la famille x86} reconnaîtra ce que l' instruction machine} \\
\levC{Simplification of meaning} & \ex{\m{Le principal du collège M. Desdouets --> Un de ses professeurs} dit de lui} \\
\levC{Change of point of view} & \ex{il présente sa démission le 18 janvier 2007 \m{suite à un débordement télévisé inconvenant envers la candidate socialiste et --> après avoir lancé une plaisanterie sur} François Hollande} \\
\levC{Questionable correction} & \ex{Des opérations de base sont disponibles dans \m{tous les --> la plupart des} jeux d' instructions} \\
\levC{Unquestionable correction} & \ex{textes \m{de René Goscinny illustrés par --> et illustrations} Albert Uderzo} \\ \hline

\levB{Spam} & \\
\levC{Obvious agrammatical spamming} & \ex{Süleyman Ier s' \m{empare de l' Arabie et fait entrer dans l' --> emp kikoo c moi ca va loll '} empire ottoman Médine et La Mecque} \\
\levC{Obvious grammatical spamming} & \ex{pour promouvoir la justice , la solidarité et \m{la paix --> l'apéro} dans le monde} \\
\levC{Subtle grammatical spamming} & \ex{Inquiété par \m{le gouvernement de Vichy --> la montée des prix du sucre}, Breton se réfugie en 1941 en Amérique} \\\hline

\end{tabular}
\end{small}
\caption{Typology of the substitutions found in the data built from the French Wikipedia}
\label{figure:typology}
\end{center}
\end{figure*}

\section{Exploiting the Data}

In this section, we illustrate some possible uses of the WiCoPaCo corpus by
describing ongoing works on spelling error corrections and paraphrase
generation selection.

\subsection{Spelling error correction}

The WiCoPaCo corpus can be used to easily build a corpus of spelling
errors.  Indeed, it can be assumed that most minor edits in documents
(i.e. edits that only concern a few words) represent orthographic,
grammatical or typographic corrections. As both the misspelled word
and its correction are available\footnote{This is assuming that the
  last revision is the correct one. To make sure that no incorrect
  corrections are kept, one can keep modification pairs $A \rightarrow
  B$ which are significantly more frequent that the reverse
  modification pairs $B \rightarrow A$ (the latter may in fact never
  occur, in particular for most spelling error corrections). One may also
  consider exploiting metadata on the user responsible for the
  modification.}, spelling errors can also be easily classified either
as \textit{non-word errors} that results in a non valid word
(e.g. when ``from'' is spelled ``rfom'') or \textit{real-word errors}
in which a correctly spelled word is substituted for another word
(e.g. ``from'' is spelled ``form'').

\paragraph*{Building a Spelling Error Corpus} So far, we have considered
editions that are limited to a single word, as most of the works on spell
checking only deal with errors at that level. We have also discarded all
editions that involve either a punctuation sign, a digit, a word with more
than one uppercase letter\footnote{Manual inspection of the corpus shows that words
with more than one uppercased letter are mostly acronyms.} or a number written in
letters\footnote{Most of the editions that involve such numbers are semantic
corrections, except when the number is \textit{une} or \textit{un} (\emph{one}).}.

First, we used a spell checker\footnote{We used the open source
  \texttt{hunspell} spell-checker
  (available from \url{http://hunspell.sourceforge.net}). In all our experiments we
  used the version 1.2.8 of \texttt{hunspell} with the version 3.4.1
  of the French dictionary \textit{Classique et r\'eforme 90}.} to
detect whether the word involved in the edition (the \textit{before
  word}) and the word that results from the edition (the \textit{after
  word}) are erroneous or not with respect to the lexicon of the spell
checker. This allows us to distinguish three kinds of editions:
\begin{itemize}
\item {\bf non-word corrections}, when the before word is erroneous and the
  after word is correct;
\item {\bf real-word error corrections and reformulations}, when both the
  before and after word are correct;
\item {\bf proper noun or foreign word editions, spam insertion and wrong
  error corrections}, when the after word is erroneous (no matter what the
  before word is).
\end{itemize}
This simple step therefore allows us to identify non-word errors and
to discard some uninteresting editions (especially proper noun and
foreign words corrections). But we still have to distinguish real-word
errors from reformulations and to remove some spam.

In a second step, we used the character edit distance between the
before word and the after word to identify both spam editions and
reformulations. Indeed, it is a well-known fact \cite{kukich92error}
that most spelling errors are within a short edit distance of their
correct form. Studying a sample of the corpus corroborates this
result: it can be observed that in an edition with an edit distance
strictly greater than 3, the word is usually completely re-written and
the edition is therefore a paraphrase or a change of meaning. It also
appears that an edition with an edit distance greater than 5 generally
corresponds to various forms of spam introduction. That is why, for
the non-word error corpus (resp. the real-word error corpus), we
discarded all the editions that involve an edit-distance larger
than~5~(resp. 3).\footnote{In our experiments, we considered that all
  operations involved in an edition have a cost of 1.}

By applying these two rules, we extracted 72,493 non-word errors and 74,100
real-word errors.

\paragraph*{French Spelling Error Patterns}

The spelling error corpus we have gathered provides valuable information
regarding spelling error patterns in French. We present here the
results of our first analysis of that corpus.

Figure~\ref{tab:err_freq} shows the most frequent editions of non-word
errors. Most of them involve a diacritic: in fact, 32.39\% of non-word error
corrections consist in only adding, changing or removing an accent. Apart from
the correction of diacritic marks, most of the corrections are caused
by the absence of a repeated consonant, which is consistent with many studies on
spelling errors in French.

For real-word errors, forgetting diacritics and errors in plurals and feminine
are causing most of the editions. It is also interesting to notice that if an
edition is frequent, the opposite edition is also frequent (for instance,
adding and removing a \texttt{s} are both frequent). Another general finding
is that 46.96\% of modifications occur at word endings, supporting our
observation that many corrections involve plural or feminine marks.

\begin{figure*}
\begin{minipage}{.45\textwidth}
  \begin{center}
  \textbf{Non-Word Errors}
  \begin{tabular}{lr||lr}
    \texttt{e \ra \'e}  & 6.7\% & \texttt{-l}        & 1.9\%\\
    \texttt{E \ra \'E}  & 6.7\% & \texttt{+i}        & 1.9\%\\
    \texttt{oe \ra \oe} & 4.6\% & \texttt{a \ra \^a} & 1.8\%\\
    \texttt{+n}         & 4.3\% & \texttt{-e}        & 1.7\%\\
    \texttt{+s}         & 2.8\% & \texttt{-n}        & 1.7\%\\
    \texttt{+r}         & 2.7\% & \texttt{+t}        & 1.6\%\\
    \texttt{\'e \ra \`e}& 2.7\% & \texttt{+m}        & 1.6\%\\
    \texttt{-s}         & 2.5\% & \texttt{e \ra \`e} & 1.4\%\\
    \texttt{+e}         & 2.2\% & \texttt{+l}        & 1.3\%\\
    \texttt{\'e \ra e}  & 2.1\% & \texttt{-r}        & 1.3\%
  \end{tabular}
  \end{center}
\end{minipage}
\begin{minipage}{.45\textwidth}
  \begin{center}
  \textbf{Real-Word Errors}
  \begin{tabular}{lr||lr}
    \texttt{+s}       & 16.2\% & \texttt{-t}          & 1.5\% \\
    \texttt{+e}       &  9.9\% & \texttt{e \ra a}     & 1.4\% \\
    \texttt{-s}       &  8.8\% & \texttt{\'e \ra er}  & 1.0\% \\
    \texttt{A \ra \`A}&  5.6\% & \texttt{er  \ra \'e} & 0.9\% \\
    \texttt{-e}       &  4.9\% & \texttt{u   \ra \`u} & 0.9\% \\
    \texttt{i \ra \^i}&  2.7\% & \texttt{\`a \ra a}   & 0.9\% \\
    \texttt{a \ra \`a}&  2.2\% & \texttt{e   \ra \'e} & 0.8\% \\
    \texttt{+nt}      &  1.9\% & \texttt{\'e \ra \`e} & 0.7\% \\
    \texttt{+t}       &  1.7\% & \texttt{s   \ra t}   & 0.7\% \\
    \texttt{a \ra e}  &  1.5\% & \texttt{\^u \ra u}   & 0.7\% 
  \end{tabular}
  \end{center}
\end{minipage}
\centering
\caption{The 20 most frequent corrections. These corrections represent
65.0\% of real-word and 53.5\% of non-word errors \label{tab:err_freq}}
\end{figure*}

\paragraph*{Spell Checker Evaluation}

Error corrections boils down to two subtasks: \textit{i)} building the set of
potential corrections of a given word (the \textit{candidate set}) and
\textit{ii)} choosing the best correction among them. In most existing
commercial or public spell-checkers, the decision of the correction to perform
is left to the user rather than performed automatically. That is why
we only only report here the evaluation of the quality of candidate sets, by counting the number of
times the correct correction is in the candidate set. Three different candidate sets
were considered:
\begin{enumerate}
\item The suggestion list of \texttt{hunspell} (denoted ``hunspell'' in the
  following). This list is built using a set of handcrafted rules that
  describe frequent error corrections and possible affixes. 
\item A list of words built by applying the most frequent edition
  scripts to the word to correct (denoted ``patterns''). This list is
  built by considering the most frequent error patterns in the corpus
  (see Table~\ref{tab:err_freq}).
\item A list of spelling error corrections extracted from the corpus
  (denoted ``pattern''). This
  list is built by gathering for each example in the train set the
  misspelled word and its correction.  
\end{enumerate}
To evaluate these approaches we randomly split our real-word
error and non-word error corpora in a training set (80\% of the
examples) and a test set (20\% of the examples). The training set is used to build the
list of corrections; the test set is used to measure the different scores.

Table~\ref{tab:cs_hit} presents the results for the three methods on the test set.
Results clearly show that the combination of the three approaches to build the candidate set
almost always produces a set that contains the correct spelling.
\begin{table}[htbp]
  \begin{tabular}{c|cc|cc}
	         & \multicolumn{2}{c|}{non-word error} & \multicolumn{2}{c}{real-word error} \\
			 & \#sugg. & corr.  & \#sugg. & corr. \\ \hline
	hunspell & 4.5     & 95.0\% & 8.6     & 65.1\% \\
	list     & 1.3     & 58.7\% & 8.3     & 75.7\% \\
	pattern  & 1.7     & 48.7\% & 2.3     & 53.2\% \\ \hline
	combi.   & 4.7     & 96.8\% & 14.9    & 92.6\%
  \end{tabular}
  \centering
  \caption{Percentage of errors for which the correct spelling is in the
    candidate set and average number of suggestions\label{tab:cs_hit}}
\end{table}

\subsection{Paraphrase Generation Selection}

Automatic paraphrasing of phrases is an active field of research, with
applications in such diverse areas as text compression, information
extraction or authoring aids. When new text is produced for human
readers, it is particularly important to ensure that the resulting
text is, in addition to semantically equivalent to the original
text\footnote{This requirement depends, of course, on the intended
  application.}, grammatical, that is as if produced by a human. One
way of ensuring local grammaticality is by reranking candidate
paraphrases in context using a language model~\cite{Bannard_05_ACL},
or by ensuring syntactic dependency conservation between the
paraphrases and their context~\cite{Max_08_GoTAL}. However, the first
approach has been shown to select more semantically incorrect
paraphrases, while the second approach takes a very conservative view
on paraphrasing. \cite{CallisonBurch_08_EMNLP} recently proposed to
condition the probability of paraphrases on the syntactic context of
an original phrase. But these probabilities have to be estimated
indirectly via pivoting in other languages, as no large high quality
representative corpora of phrasal paraphrases exist to derive valid
rewriting syntactic patterns.

\begin{figure*}[!htpb]
\begin{center}
\begin{small}
\begin{tabular}{|c|c||c|c||c|c||c|c|}
\hline
\multicolumn{2}{|c||}{pos$_{1}$: ADJ} & \multicolumn{2}{|c||}{pos$_{1}$: ADJ ADJ} & \multicolumn{2}{|c||}{pos$_{1}$: DET ADJ NOM} & \multicolumn{2}{|c||}{pos$_{1}$: VER PRP DET NOM} \\ \hline
pos$_{2}$ & $p($pos$_{2}|$pos$_{1})$ & pos$_{2}$ & $p($pos$_{2}|$pos$_{1})$ & pos$_{2}$ & $p($pos$_{2}|$pos$_{1})$ & pos$_{2}$ & $p($pos$_{2}|$pos$_{1})$ \\ \hline \hline
ADJ & 0.4029 & ADJ & 0.2371 & DET NOM & 0.3081 & VER & 0.1666 \\
NOM & 0.1221 & NOM & 0.1666 & DET ADJ NOM & 0.0880 & VER DET NOM & 0.0555 \\
VER & 0.1116 & NOM ADJ & 0.0576 & VER & 0.0314 & ADJ & 0.0555 \\
PRP NOM & 0.0350 & VER & 0.0448 & DET NOM ADJ & 0.0314 & VER VER & 0.0476 \\
NOM ADJ & 0.0156 & ADJ ADJ & 0.0384 & PRO & 0.0251 & DET NOM & 0.0396 \\
ADV ADJ & 0.0126 & ADJ PUN & 0.0256 & PRP NOM & 0.0251 & VER PRP NOM & 0.0317 \\ \hline
\end{tabular}
\end{small}
\caption{Distribution of part-of-speech sequences for rewritings for French phrases}
\label{figure:pos}
\end{center}
\end{figure*}

A stumbling block for research on paraphrasing and rewriting in
general is the lack of available corpora for learning models and
assessing their performance on naturally-occurring data. To our
knowledge, to date, no large resources of such rewritings have been
made available. Most corpora are built with the aim to support
specific research projects. For example, for their study on
abstractive sentence compression~\cite{Cohn_08_COLING}, the authors
artificially created their own corpus from 575 sentences. Other
corpora, such as the Ziff-Davis corpus~\cite{Knight_02_AI}, built from
pairs of documents and abstracts, mostly focus on a single phenomena
(e.g. word deletion in 1067 sentence pairs).

As for spelling corrections, our corpus of modifications mined from
Wikipedia's revision history can be used to increase by an order of
magnitude the quantity of available data. Furthermore, the fact that
these data can be naturally-occurring pairs of rewritings is certainly
the most important characteristic. Lastly, if data can be correctly
classified, held-out data can be trivially extracted (and possibly
validated by humans) to be used as evaluation sets.

This type of corpus was already used for the specific task of text
compression~\cite{Nelken_08_AAAI}, where the authors report using a
set of 380,000 pairs of full and compressed sentences extracted
from a subset of the English Wikipedia. However, with no automatic
classification of edits, they made the simplifying assumption that all
observed text compression preserved meaning to build large lexicalized
models for the task. 


An interesting characteristic of our resource is that it
allows associating valid \emph{grammatical} rewritings, either at the lexical,
morpho-syntactic, or syntactic level. We make the hypothesis that for local
paraphrase generation all data from our resource can be exploited,
independently of the fact that some rewritings introduce some meaning change
or not.\footnote{This is assuming that other models for meaning
  conservation are also used when assessing automatic paraphrases.} For example, models of morpho-syntactic rewriting patterns can
be built from a subset of the resource to build grammatical models
based on valid morpho-syntactic patterns. An illustration is given on
Figure~\ref{figure:pos}, which provides some examples of the distribution of
part-of-speech sequences for rewritings derived from our
resource.\footnote{To build those patterns, the \texttt{TreeTagger}
  analyzer (available from
  \url{http://www.ims.uni-stuttgart.de/projekte/corplex/TreeTagger/})
  was used.} When several candidate paraphrases are produced
automatically by a generative method, information such as a sequence
of two adjectives (ADJ ADJ) has a 0.2371 probability of being
rewritten as a single adjective (ADJ) can be effectively exploited to
assess the grammaticality of such rewritings. If the observed
distribution of rewriting may adequately reflect natural rewriting
patterns, further studies may also show some bias due to the genre of
texts being rewritten and the types of contributors.

\section{Conclusions and future work}

In this article, we have introduced the freely available WiCoPaCo corpus of rewritings
in context automatically extracted from Wikipedia's revision
history. It is, to our knowledge, one of the largest corpus of
naturally-occurring rewritings, which can be exploited, as was shown
with our experiments on spelling errors and morpho-syntactic rewriting
patterns, on many levels. 

Our future work will be on two areas. First, we want to carry out a
more detailed analysis of the types of rewriting found in the resource
to identify classes that may be of use for specific studies or
applications. Automatic classification will be the next step, as the
ability to classify rewritings will be needed for training algorithms
or applications such as linguistic-aware text authoring or spam
reporting. 

This type of work exploit latent data at very little cost (some CPU time and disk space) than can be very useful for linguistic
studies and NLP applications. We therefore also plan to reproduce our
methodology on other languages\footnote{Wikipedia is available in as
  many as 269 languages as of March, 2010, and our techniques can be
  applied on all languages with mostly clear word segmentation.} and
other wikis such as WikiNews for news articles.

\section*{Acknowledgements} 

This work was funded by a LIMSI AI grant and the ANR Trace project. The authors wish to thank Julien
Boulet and Martine Hurault-Plantet for their contribution to this work.

\bibliographystyle{lrec2006}
\bibliography{biblio}

\end{document}